\documentclass[letterpaper, 10 pt, journal, twoside]{IEEEtran}
\IEEEoverridecommandlockouts

\usepackage{verbatim}
\usepackage[pdftex]{graphicx}
\usepackage{epsfig}
\usepackage{amsmath,amssymb,amsfonts,bm,amscd} 
\usepackage{mathrsfs}
\usepackage{setspace}
\usepackage{fancyhdr}
\usepackage{balance}

\usepackage{psfrag}
\usepackage{cite}
\usepackage{tikz}
\usepackage[utf8]{inputenc}
\usepackage[ruled,vlined,linesnumbered,noresetcount]{algorithm2e}
\usepackage{pgfplots} 
\usepackage{pgfgantt}
\usepackage{pdflscape}
\usepackage[short]{optidef}
\pgfplotsset{compat=newest} 
\pgfplotsset{plot coordinates/math parser=false}
\usepackage[hidelinks]{hyperref}
\usepackage{stmaryrd}
\usepackage{booktabs}
\usepackage{multirow}
\usepackage{siunitx}

\usepackage[caption=false,font=footnotesize]{subfig}

\newlength\fwidth

\newcommand{\norm}[1]{\left\Vert#1\right\Vert}

\newcommand{\RR}{{\mathbb R}}

\newcommand{\bhat}[1]{\mathbf{\hat{\text{$#1$}}}}

\usepackage{graphics}

\title{
Online Trajectory Generation with Distributed Model Predictive Control for Multi-Robot Motion Planning
}

\author{Carlos E. Luis$^{1}$, Marijan Vukosavljev$^{1}$ and Angela P. Schoellig$^{1}$%
	\thanks{Manuscript received: September 10, 2019; Revised November 27, 2019; Accepted December 18, 2019.}
	\thanks{This paper was recommended for publication by Editor Nak Young Chong upon evaluation of the Associate Editor and Reviewers' comments. This work was supported by NSERC research and equipment grants (RTI 2018-00847, CRDPJ 528161-18, CREATE 466088), and the CFI JELF/ORF grant \#33000.} 
	\thanks{$^{1}$Carlos E. Luis, Marijan Vukosavljev and Angela P. Schoellig are with the Dynamic Systems Lab \href{www.dynsyslab.org}{(www.dynsyslab.org)}, Institute for Aerospace Studies, University of Toronto, Canada. E-mails:
	\{{\tt\footnotesize carlos.luis, mario.vukosavljev, angela.schoellig}\}	{\tt\footnotesize @robotics.utias.utoronto.ca}}%
	\thanks{Digital Object Identifier (DOI): see top of this page.}
}

\usepackage{xcolor,colortbl}

\newcommand{\todo}[1]{\textcolor{black}{#1}}

\usepackage{bm}

\begin{document}
\markboth{IEEE Robotics and Automation Letters. Preprint Version. Accepted December, 2019}
{Luis \MakeLowercase{\textit{et al.}}: Online Trajectory Generation for Multi-Robot Motion Planning} 

\setlength{\parskip}{0pt}
\maketitle

\begin{abstract}
    We present a distributed model predictive control (DMPC) algorithm to generate trajectories in real-time for multiple robots. We adopted the \textit{on-demand collision avoidance} method presented in previous work to efficiently compute non-colliding trajectories in transition tasks. An event-triggered replanning strategy is proposed to account for disturbances. Our simulation results show that the proposed collision avoidance method can reduce, on average, around 50\% of the travel time required to complete a multi-agent point-to-point transition when compared to the well-studied Buffered Voronoi Cells (BVC) approach. Additionally, it shows a higher success rate in transition tasks with a high density of agents, with more than 90\% success rate with 30 palm-sized quadrotor agents in a 18~$\text{m}^3$ arena. The approach was experimentally validated with a swarm of up to 20 drones flying in close proximity.
\end{abstract}



\setcounter{section}{0}
\section{Introduction}
\label{sec:introduction}

\IEEEPARstart{O}{nline} trajectory generation is key to execute missions in dynamic or unknown environments. In particular, multi-robot tasks are especially challenging due to a high number of decision-making agents sharing the same space. In such settings, the planning algorithms must compute collision-free and goal-oriented trajectories, taking into account the state of the environment and neighbouring agents.

A wide variety of techniques exist to tackle the multi-robot trajectory generation problem. First, optimization-based techniques such as Sequential Convex Programming (SCP) \cite{augugliaro2012generation, chen2015decoupled} and Distributed Model Predictive Control (DMPC) \cite{van2017distributed, luis2019trajectory} have successfully solved point-to-point trajectory generation problems for multiple agents. Second, discrete planning strategies such as Rapidly-exploring Random Trees (RRT)~\cite{vcap2013multi} have been extended to the multi-agent case. Third, a combination of discrete planning and continuous optimization has been developed to coordinate multiple robots in cluttered environments \cite{honig2018trajectory}. GPU-accelerated approaches can reduce the runtime of these offline planners \cite{hamer2018fast}.

Real-time trajectory generation is required for quick adaptation in dynamic environments, but it remains challenging to implement for robot swarms. Optimal Reciprocal Collision Avoidance (ORCA) and all its variants have pushed towards real-time trajectory generation~\cite{van2011reciprocal}, providing experimental validation with various robotic platforms in planar environments \cite{alonso2018cooperative}. A similar approach achieves collision avoidance through the concept of Buffered Voronoi Cells (BVC) \cite{zhou2017fast}, showing initial results of online trajectory generation in 2D with multiple quadrotors operating at a fixed height. The BVC concept has been recently used in tandem with discrete planners \cite{csenbacslar2019robust}, primarily to avoid deadlocks in scenarios where plain BVC would get trapped and fail the task.

Robust MPC frameworks such as tube MPC have been developed for distributed multi-agent systems under uncertainty, both with linear \cite{hernandez2016distributed} and nonlinear \cite{nikou2018decentralized} dynamics. Although both approaches provide proofs and simulation results, they are not real-time implementable with current hardware and solver capabilities. 


\begin{figure}[t]
	\centering
	\includegraphics[width=0.8\columnwidth]{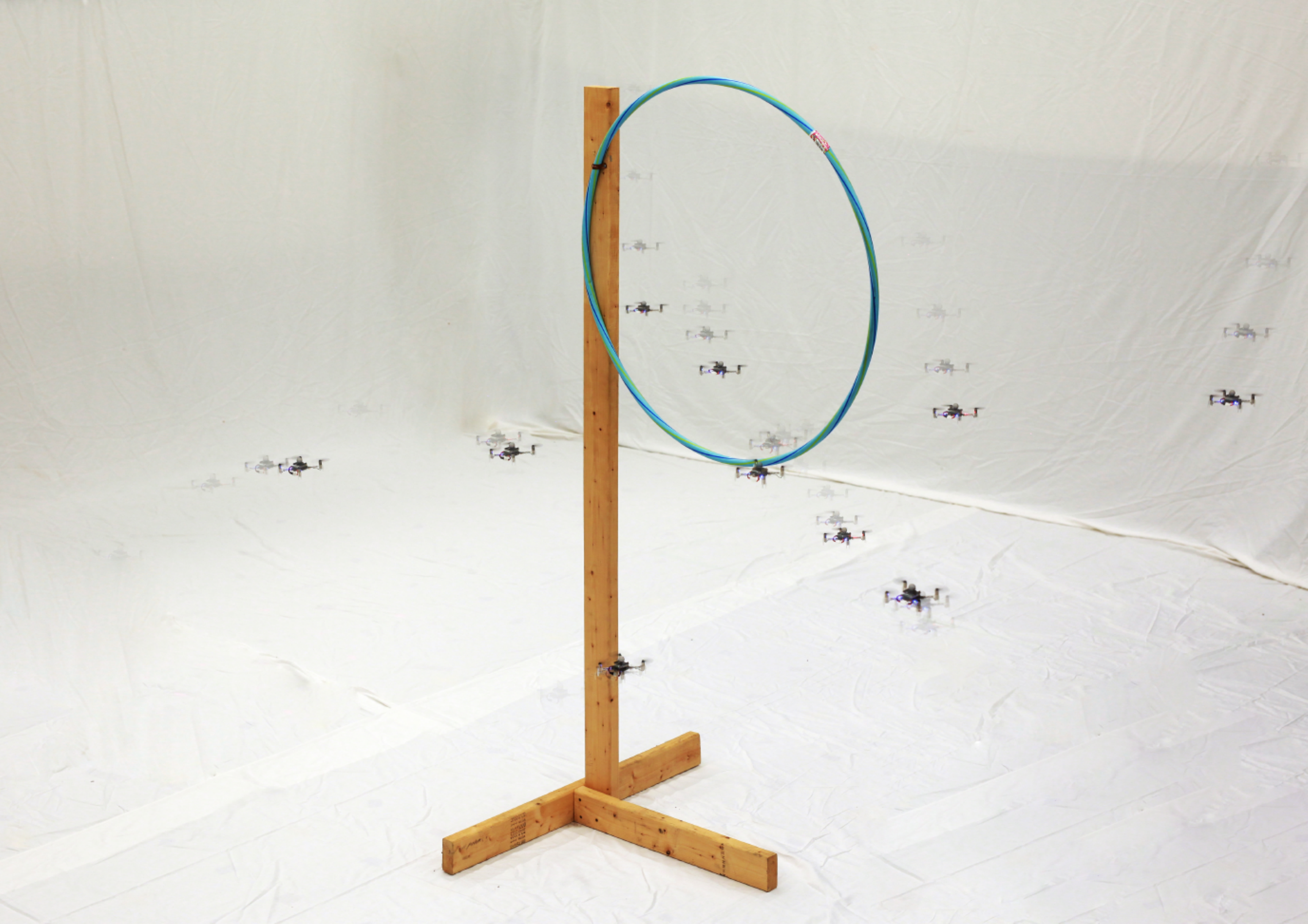}
	\caption{A ten-drone transition task through a hula-hoop solved using our proposed online trajectory generation method. Our distributed computation allows for real-time multi-robot motion planning, enabling complex transition tasks to be performed. A video of the performance is found at 
		\href{http://tiny.cc/online-dmpc}{{\tt http://tiny.cc/online-dmpc}}.}
	\label{fig:avoidance}
\end{figure}

\todo{We present a novel real-time, multi-vehicle motion planning framework that significantly outperforms existing methods in terms of the success rate to complete transition tasks in agent-dense environments. To the best of our knowledge, this paper presents the first results on real-time motion planning for drone swarms of up to 20 drones, executed from a single off-board computer. The proposed algorithm is implemented in a centralized fashion, and relies on information sharing between agents. We developed a DMPC formulation of the problem that includes state feedback, which enables online replanning and therefore increases overall robustness.} As such, our framework provides an essential functionality for higher-level planners that specify complex team missions in terms of goal locations to be visited by the agents. Compared to our offline approach \cite{luis2019trajectory}, our main contributions are threefold: (\textit{i}) a multi-agent motion planning framework based on distributed model predictive control, which allows for real-time trajectory generation, (\textit{ii}) an event-triggered replanning strategy for robust execution of plans and (\textit{iii}) a thorough empirical evaluation of the method. 

Our approach contrasts from current online methods (e.g.,~\cite{csenbacslar2019robust}) in that:
\begin{itemize}
    \item It is purely optimization-based, in the form of a standard and efficiently-solvable \todo{Quadratic Program (QP)}.
    \item It uses \textit{on-demand collision avoidance} instead of the BVC method for partitioning the free space, resulting in less conservative movement and faster transition times. 
\end{itemize}


The rest of the paper is organized as follows: Sec. II introduces the problem. Sec. III formalizes the DMPC method and Sec.~IV introduces the trajectory replanning strategy. The algorithm for input updates is presented in Sec.~V. Finally, Sec.~VI and VII provide simulation and experimental results of our approach with teams of drones.
\section{Problem Statement}
\label{sec:approach}
Given $N$ agents with known linear dynamics, a finite 3-dimensional workspace $\mathcal{W} \subset \RR^3$, desired end positions $\textbf{p}_{d,i} \in \mathcal{W}$ for each agent $i$ and static obstacle set $\mathcal{E} \subset \mathcal{W}$, compute inputs $\textbf{u}_i[k] \in \RR^3$ for each agent such that:
\begin{itemize}
	\item the agents do not collide with each other or with the obstacles;
	\item the agents remain within $\mathcal{W}$ for all time;
	\item there exists a time $T_f$ after which the agents remain sufficiently close to their desired positions.
\end{itemize}

\subsection{The Agents}
\begin{figure}[t]
	\centering
	\includegraphics[width = \columnwidth]{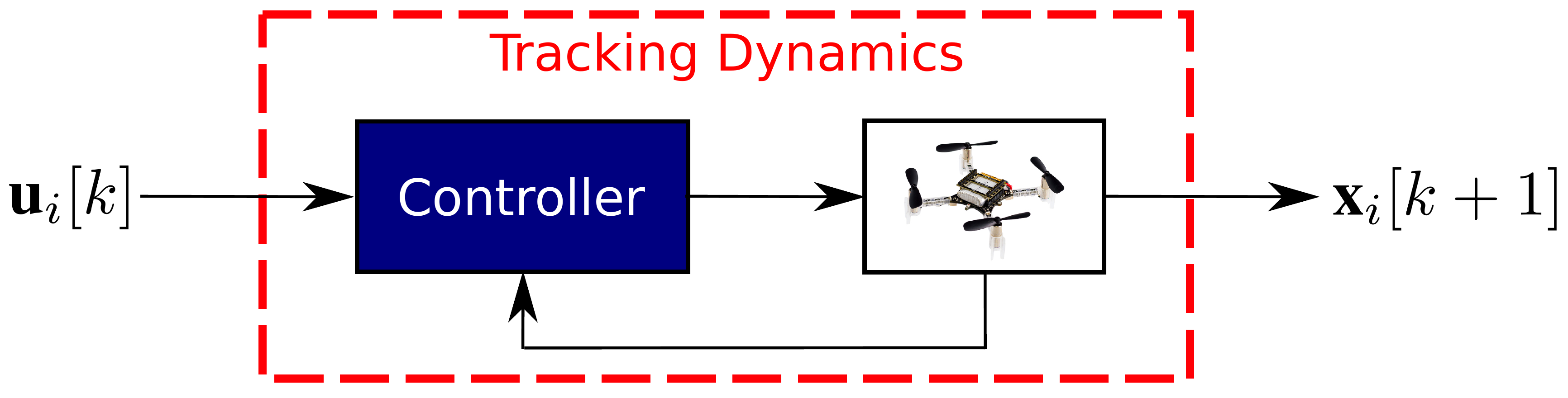}
	\caption{Block diagram of the control system of agent $i$. Here we depict the agent as a Crazyflie 2.0 quadrotor, which is our experimental platform.}
	\label{fig:block}
\end{figure}

We assume every agent $i$ is equipped with a controller for position trajectory tracking and $\textbf{u}_i$ is a position reference, as shown in Fig.~\ref{fig:block}.

Furthermore, assume each agent $i$ obeys some known trajectory tracking dynamics given by a discrete linear system:
\begin{equation}
\label{eq:sys}
    \textbf{x}_i[k + 1] = \textbf{A}_i\textbf{x}_i[k] + \textbf{B}_i\textbf{u}_i[k].
\end{equation}

For example, in this paper we consider the system~(\ref{eq:sys}) to represent a quadrotor with an underlying position controller \cite{mellinger2011minimum}, for which the input ($\textbf{u}_i[k] \in \RR^3$) is a position reference signal, and the states ($\textbf{x}_i[k] \in \RR^6$) are the position and velocity of the vehicle, i.e., ${\textbf{x}_i[k] = (\textbf{p}_i[k], \, \textbf{v}_i[k])}$. This results in a second-order system defining the dynamics, with  $\textbf{A}_i \in \RR^{6 \times 6}$, $\textbf{B}_i \in \RR^{6 \times 3}$. \todo{One may adopt more complex inputs (e.g., adding a velocity reference), or more complex systems (e.g., other differentially flat robots) as long as the dynamics can be represented by a linear system.}

\section{Online Distributed Model Predictive Control (DMPC)}
\label{sec:dmpc}
In this section we formalize the MPC optimization problem solved in real-time for each agent. The approach is based on the offline method presented in \cite{luis2019trajectory}.


\begin{figure*}
	\centering
	\hfill
	\subfloat[Colliding scenario]{\label{fig:problem}\includegraphics[width=0.32\textwidth]{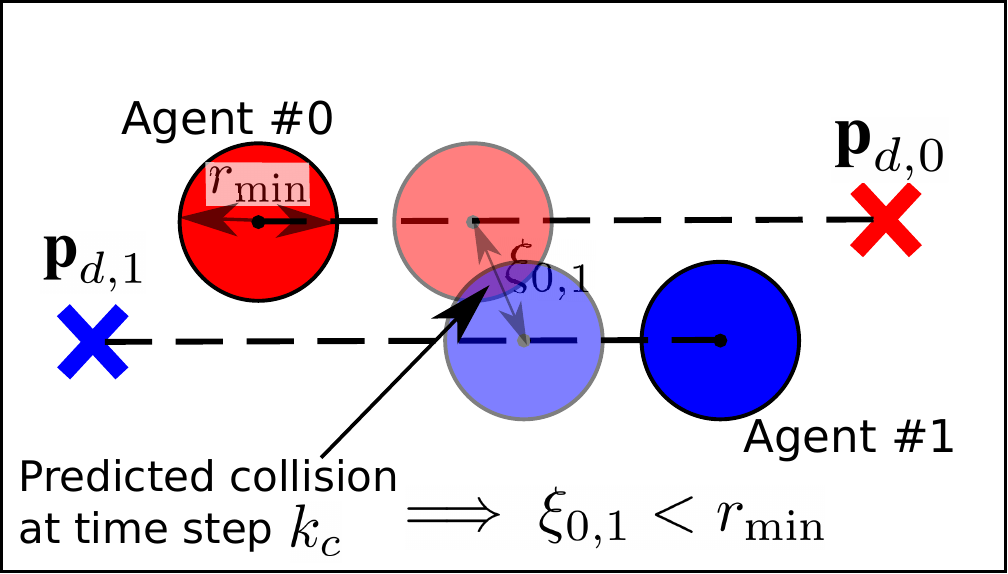}}
	\hspace{1ex}
	\subfloat[BVC collision avoidance]{\label{fig:bvc}\includegraphics[width=0.32\textwidth]{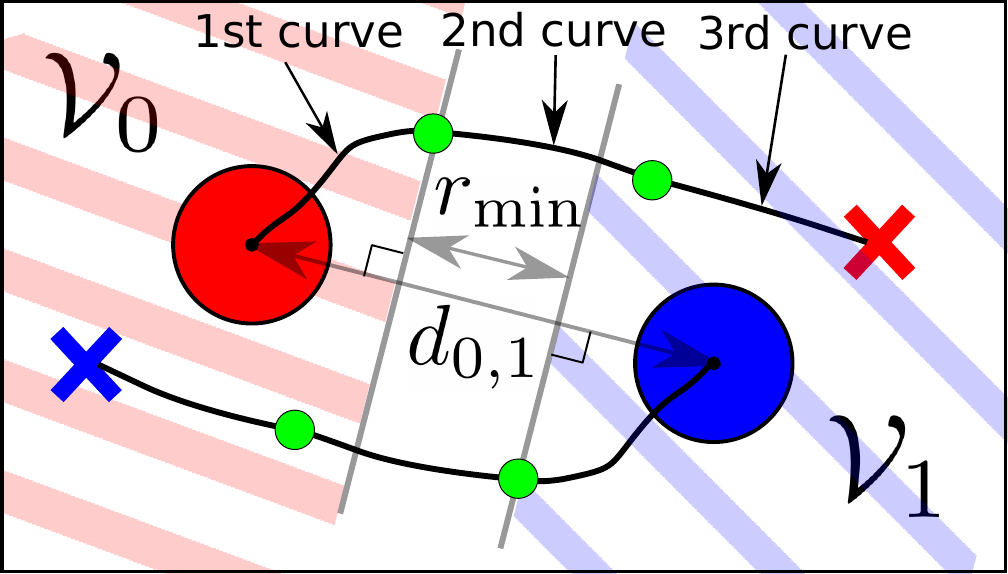}}
	\hspace{1ex}
	\subfloat[On-demand collision avoidance]{\label{fig:ondemand}\includegraphics[width=0.32\textwidth]{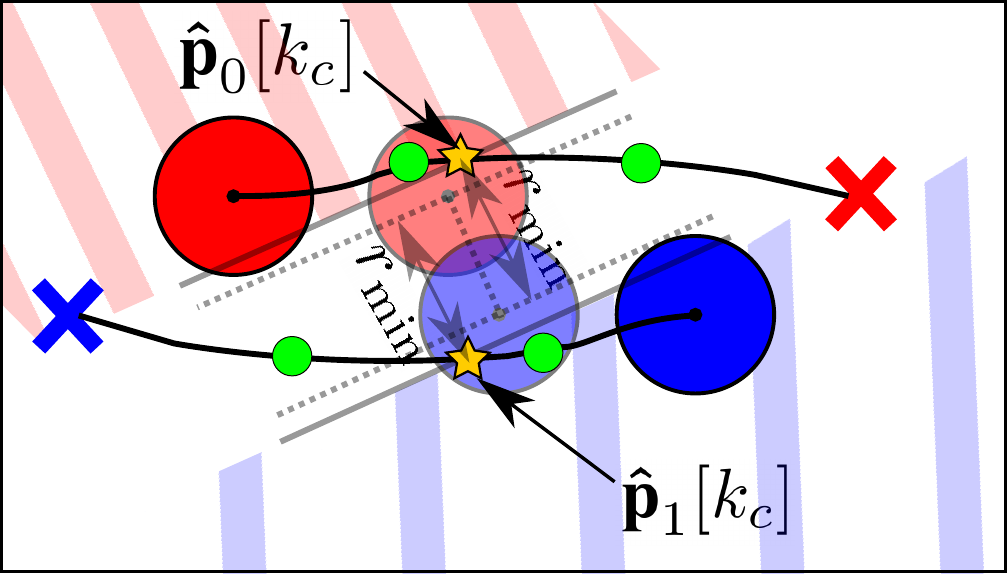}}
	\caption{Two-agent transition scenario in 2D. The agents are represented by a circle of certain radius. The X marks the intended goal of each agent. In \protect\subref{fig:problem} the dashed lines represent the nominal (colliding) trajectories, where the translucent circles represent the position of each agent at time step $k_{c}$ in which the first collision is predicted. In \protect\subref{fig:bvc} we show the input update using the BVC method. The green dots represent the concatenation points of the Bézier curves. The first \todo{curve} is constrained to lie within the coloured zone for each agent. In \protect\subref{fig:ondemand} the agents update their inputs using on-demand collision avoidance, leveraging the predicted collision information to build the separating hyperplanes. The star represents the sample of the updated input which was constrained to be within the coloured zone.}
	\label{fig:collision_avoidance}
\end{figure*}

\subsection{Trajectory Parameterization}
\label{subsec:input}
Our approach is based on receding horizon control, meaning that at the discrete time step $k_t$, corresponding to the continuous time instant $t_0$, we recompute the input sequence to be applied over a finite horizon of $K$ time steps. Given a desired discrete time step duration $h$, we get the continuous time horizon duration $t_h = (K-1)h$. We parameterize the continuous input signal $\textbf{u}_i(t)$ for $t \in [t_0,\, t_0+t_h]$ as a concatenation of $l$ Bézier curves, similar to \cite{honig2018trajectory}. For more details on Bézier curves we refer the reader to \cite{joy2000bernstein}.

We select Bézier curves since we can impose smoothness requirements in the input and can easily represent its derivatives. In order to define a Bézier curve in $\RR^3$ of arbitrary degree $p$ and duration $T$, first we must construct the $p+1$ Bernstein polynomials of degree $p$:
\begin{equation}
\todo{S_{m,p}(t)} = \begin{pmatrix}
p \\
m
\end{pmatrix} (1-t/T)^{p-m}(t/T)^m \quad \forall t \in (0,\, T),
\label{eq:bernie}
\end{equation}
with $m = 0,1,\dots,p$. Now, an $n$-dimensional Bézier curve of degree $p$ is defined as $\textbf{\todo{S}}(t) = \sum_{i=0}^{p}\textbf{P}_m \todo{S_{m,p}(t)}$
with $\textbf{P}_m \in \RR^3$. The set $\mathcal{P} = \lbrace \textbf{P}_0, \textbf{P}_1, \dots, \textbf{P}_p \rbrace$ represents the ${p+1}$ control points that uniquely characterize the curve. The control points are a finite parameterization of the continuous curve and serve as the optimization variables to compute the agents' trajectories over the horizon. 

\todo{Samples of $\textbf{\todo{S}}(t)$ and its derivatives can be computed as a linear combination of its control points \cite{joy2000bernstein}}, which will be used in what follows to build a convex optimization problem.

\subsection{The Agent Prediction Model}

We introduce the notation $\bhat{(\cdot)}[k|k_t]$, which represents the predicted value of $(\cdot)[k_t+k]$ with the information available at $k_t$ and $k \in \lbrace 0, \dots, K-1 \rbrace$, where $K$ is the horizon length. The prediction model of agent $i$ is given by
\begin{equation}
\label{eq:prediction}
\bhat{\textbf{x}}_i[k + 1|k_t] = \textbf{A}_i\bhat{\textbf{x}}_i[k|k_t] + \textbf{B}_i\bhat{\textbf{u}}_i[k|k_t].
\end{equation}

Using (\ref{eq:prediction}) we can represent the (stacked) predicted state sequence over the horizon, $\textbf{X}_i \in \RR^{6K}$, as 
\begin{equation}
\label{eq:states}
\textbf{X}_i = \textbf{A}_{0,i} \bar{\textbf{x}}_i[k_t] + \bm{\Lambda}_i\textbf{U}_i,
\end{equation}
where $\textbf{U}_i \in \RR^{3K}$ is the stacked input sequence,  $\bar{\textbf{x}}_i[k_t]$ is the measured state at $k_t$, and matrices $\textbf{A}_{0,i} \in \RR^{6K \times 6}$ and $\bm{\Lambda}_i \in \RR^{3K \times 3K}$. We note that $\textbf{U}_i$ is a \textit{sampled} representation of the input, and it can be obtained from a linear combination of the control points of a \textit{continuous} Bézier curve. We define $\bm{\mathcal{U}}_i \in \RR^{3l(p+1)}$ as the decision vector to optimize, which represents the control points of the $l$ Bézier curves of degree~$p$.

\subsection{Input Continuity}
Trajectory smoothness is enforced through equality constraints. First, the initial control point of the input is chosen to be equal to a constant vector; the way this constant vector is constructed is the subject of Sec.~\ref{subsec:replan}. Second, continuity between the $l$ Bézier curves is guaranteed up to a certain derivative by forcing the endpoint of a curve to match the beginning of the next curve, i.e., the difference between control points must be equal to zero \cite{csenbacslar2019robust}.

Using linear relationships between the control points of the Bézier curve and the control points of its derivatives, we build a tuple $(\textbf{A}_{\text{eq}}, \textbf{b}_{\text{eq}})$ that represents the input continuity constraints of the form $\textbf{A}_{\text{eq}}\bm{\mathcal{U}}_i =  \textbf{b}_{\text{eq}}$ for each agent $i$. 

\subsection{\todo{Physical Limits of the Robot}}
Since the agents have limited actuation and the workspace is limited as defined by $\mathcal{W}$, we must encode such limitations within the optimization. For dynamic feasibility we impose the following constraints
\begin{equation}
\bm{\gamma}^{(c)}_{\min} \leq \frac{d^c}{dt^c}\textbf{u}_i(t) \leq \bm{\gamma}^{(c)}_{\max}, \quad c = \lbrace 0, 1, \dots, r \rbrace,
\end{equation}
where $\bm{\gamma}^{(c)}_{\min}$ and $\bm{\gamma}^{(c)}_{\max}$ are the given maximum and minimum values of the $c^{th}$ derivative of the input.

In general, imposing these constraints has posed difficulties in past work. One option proposed in the literature is to exploit the convex hull property of Bézier curves, although this may impose overly conservative bounds \cite{mercy2017spline}. A second option, as suggested in \cite{csenbacslar2019robust}, is to not impose the constraints at all and check afterwards if the trajectories comply with the constraints; if not, the problem needs to be resolved for constraint satisfaction. In this work we propose a third alternative, in which we obtain specific samples of the input and its derivatives (as a linear combination of the control points) and limit those appropriately through linear inequality constraints of the form $\textbf{A}_{\text{ineq}}\bm{\mathcal{U}}_i =  \textbf{b}_{\text{ineq}}$. The procedure involves computing a linear transformation between control points and polynomial coefficients in the power basis \cite{joy2000bernstein}, which then can be multiplied by vectors of the form $\lbrace1,\, t_0,\, \dots, t_0^p \rbrace$ to obtain the exact value of $\textbf{\todo{S}}(t_0)$ and its derivatives. 

This method avoids the conservativeness of using the convex hull property and the potential need to resolve the problem as in \cite{csenbacslar2019robust}. \todo{The values for $\bm{\gamma}^{(c)}_{\min}$ and $\bm{\gamma}^{(c)}_{\max}$ were picked based on experimental understanding of the specific robotic platform being used. In the case of quadrotors we found that limiting the acceleration ($c = 2$) led to good tracking performance of the underlying controller.}

\subsection{Optimization-Based Collision Avoidance}
\label{subsec: avoidance}
For collision avoidance we require the following inequality to hold throughout trajectory execution
\begin{equation}
	\norm{\bm{\Theta}^{-1} (\textbf{p}_i[k_t] - \textbf{p}_j[k_t])}_2 \geq r_{\min}, \quad \forall j \neq i,
\end{equation}
where $\bm{\Theta}$ is a scaling matrix to obtain general ellipsoid safety boundaries, and $r_{\min}$ is the minimum distance between two agents before collision.

We explored two approaches: Buffered Voronoi Cells (BVC) \cite{zhou2017fast,csenbacslar2019robust} and on-demand collision avoidance \cite{luis2019trajectory}. Both methods rely on the same principle of imposing hyperplane constraints that limit the available free space over which the agent is allowed to optimize its future inputs. In Fig.~\ref{fig:problem} we present a simple collision avoidance scenario with two agents in 2D. 


In the BVC method, the agents are restricted to remain within their own Buffered Voronoi Cell, $\mathcal{V}_i$, for a time $\tau$ of their horizon. In this work, we define a Buffered Voronoi Cell similar to \cite{zhou2017fast} but including the scaling matrix:
\begin{equation}
\label{eq: voronoi}
\resizebox{\columnwidth}{!}{%
	$\mathcal{V}_i = \left\{   \textbf{p} \in \RR^3\; \middle| \;\frac{\bm{\Theta}^{-2}(\textbf{p}_i - \textbf{p}_j)^\intercal(\textbf{p} - \textbf{p}_i)}{d_{i,j}}  \geq \frac{r_{\min} - d_{i,j}}{2} \right\}, \forall j \neq i,$%
}
\end{equation}
where $d_{i,j} = \norm{\bm{\Theta}^{-1}(\textbf{p}_i - \textbf{p}_j)}_2$, and $\textbf{p}_i$, $\textbf{p}_j$ are the measured positions of agents $i$ and $j$ at time step $k_t$. Fig.~\ref{fig:bvc} shows the BVCs calculated (shaded areas) for our two-agent example. The condition in (\ref{eq: voronoi}) defines a linear constraint on the position of the agents to achieve collision avoidance. Let $\mathcal{P}_{i,1}$ be the set of control points of agent $i$ corresponding to the first Bézier curve of the input. To achieve collision avoidance we impose the constraint ${\mathcal{P}_{i,1} \in \mathcal{V}_i}$, which translates to ${p+1}$ constraints on the control points. 


Collision-free updates are achieved with this method, as shown in Fig.~\ref{fig:bvc}.


On the other hand, the on-demand method of \cite{luis2019trajectory} relies on a predict-avoid paradigm for collision avoidance. It assumes communicative agents that share with the team a representation of their future actions. \todo{In our case, since the input and the state are closely related (reference trajectory and measured position)}, we have two options for collision avoidance:
\begin{itemize}
	\item \textbf{State space}: constraints are imposed on the predicted states $\textbf{X}_i$ of the agents, which can be obtained as a linear combination of the optimal inputs using (\ref{eq:states}). This results in collision-free \textit{predicted} positions over the horizon.
	\item \textbf{Input space:} constraints are imposed on the inputs $\textbf{U}_i$ directly, resulting in collision-free \textit{reference} positions over the horizon. 
\end{itemize}
\todo{For the general dynamics in system (\ref{eq:sys}), non-intersecting trajectories in the input space would not necessarily achieve collision avoidance}.

Agent $i$ detects the first predicted collision (in the state space) with any neighbour $j$ at time step $k_{c,i}$ whenever
\begin{equation}
\label{eqn:detect}
\xi_{ij} = \norm{\bm{\Theta}^{-1}\left( \bhat{\textbf{p}}_i[k_{c,i}|k_t-1]-\bhat{\textbf{p}}_j[k_{c,i}|k_t-1]\right) }_2 \geq r_{\min},
\end{equation}
does not hold. For input space detection it suffices to replace predicted positions with predicted inputs (position reference). We define a subset $\Omega_i$ of neighbours of agent $i$ for which collision constraints are constructed, defined as:
\begin{equation}
\Omega_i = \lbrace j \in \lbrace 1,\dots,N \rbrace \mid \xi_{ij} < g(r_{\min}), \, j \neq i \rbrace,
\end{equation}
where $g(r_{\min})$ models the area around the agent for which collision avoidance is required. In this work we used $g(r_{\min}) = 2r_{\min}$. As proposed in \cite{luis2019trajectory}, we can procure collision avoidance in the state space by enforcing a first-order approximation of the constraint
\begin{equation}
\label{eq:to_linearize}
\resizebox{\columnwidth}{!}{%
	$\norm{\bm{\Theta}^{-1}\left( \bhat{\textbf{p}}_i[k_{c,i}-1|k_t]-\bhat{\textbf{p}}_j[k_{c,i}|k_t-1]\right) }_2 \geq r_{\min} + \varepsilon_{ij}, \, \forall j \in \Omega_i,$}
\end{equation}
where $\varepsilon_{ij} < 0$ are slack decision variables for relaxation. 


Note that on-demand avoidance only constrains a specific sample of the curve at $k_{c,i}$, as shown with the yellow stars in Fig.~\ref{fig:ondemand}. This sample must lie within a partition of the space given by the linearization of (\ref{eq:to_linearize}), whereas BVC constrains a complete segment of the curve. Comparing the resulting trajectories in Fig.~\ref{fig:bvc} and Fig.~\ref{fig:ondemand}, it is clear that on-demand avoidance leads to less conservative maneuvers than the BVC method. In Sec.~\ref{sec:simulation} we analyze how these insights impact the ability to complete multi-agent transition tasks.

In both cases, to implement collision avoidance we need only add an inequality constraint tuple ($\textbf{A}_{\text{coll}},\, \textbf{b}_{\text{coll}}$) that satisfies $\textbf{A}_{\text{coll}}\bm{\mathcal{U}}_i \leq  \textbf{b}_{\text{coll}}$.



\subsection{Cost Function}
We search to minimize a cost function which results from the sum of various terms. In this section we omit the subindex $i$ for the tuning parameters of each term of the cost function, but each agent could have different values.

\subsubsection{Error to goal} this term drives the agent to its goal location. We aim to minimize the sum of errors between the positions at the last $\kappa < K$ time steps of the horizon and the goal location $\textbf{p}_{d,i}$. The quadratic cost function is defined as 
\begin{equation}
\label{eq:error_cost}
\mathcal{J}_{i,\text{error}} = \sum _{k = K - \kappa}^K q_k\norm {\bhat{\textbf{p}}_i[k|k_t]-\textbf{p}_{d,i}}_2^2,
\end{equation}
where $q_k > 0$ are the  positive weights of each time step. \todo{This expression can be formulated as a quadratic form in terms of the inputs and the measured state at $k_t$.}
\subsubsection{Energy} we minimize a weighted combination of the sum of squared derivatives, as in \cite{honig2018trajectory,charles2013polynomial}. The cost is defined as
\begin{equation}
	\mathcal{J}_{i,\text{energy}} = \sum_{c=0}^r \alpha_c \int_{0}^{t_h}\norm{\frac{d^c}{dt^c} \bhat{\textbf{u}}_i(t)}_2^2dt,
\end{equation}
where $\alpha_c > 0$ is a scalar weight for each derivative of the input, until the $r^{th}$ derivative. This term can be evaluated in closed form to get a quadratic form in terms of $\bm{\mathcal{U}}_i$ \cite{charles2013polynomial}. 
\subsubsection{Collision constraint violation} we implement on-demand collision avoidance as soft constraints, which requires a penalty term to be added in the cost function to limit the amount of relaxation of the constraints. For that we consider both quadratic and linear penalty costs
\begin{equation}
	\mathcal{J}_{i,\text{violation}} = \zeta \norm{\varepsilon_{ij}}_2^2 +  \xi\varepsilon_{ij},
\end{equation}
where $\zeta$ and $\xi$ are the weights of each term.

A similar approach can be used to relax the constraints in the BVC method, with the difference that for each neighbour $j$ we add penalty terms for the $p + 1$ constraints on the control points of the first Bézier curve segment.

All the elements previously mentioned compose the following standard QP problem:
\begin{mini}|l|
	{\bm{\mathcal{U}}_i, \varepsilon_{ij}}{\mathcal{J}_{i,\text{error}} + \mathcal{J}_{i,\text{energy}} + \mathcal{J}_{i,\text{violation}}}{}{}
	{\label{eqn:convex_coll}}{}
	\addConstraint{\textbf{A}_{\text{eq}}\bm{\mathcal{U}}_i}{= \textbf{b}_{\text{eq}}}
	\addConstraint{\textbf{A}_{\text{in}}\bm{\mathcal{U}}_i}{\leq \textbf{b}_{\text{in}}}
	\addConstraint{\textbf{A}_{\text{coll}}\bm{\mathcal{U}}_i}{\leq \textbf{b}_{\text{coll}}}
	\addConstraint{\varepsilon_{ij}}{\leq 0 \quad \forall j \in \Omega_i}.
\end{mini}
\section{Event-triggered Replanning}
\label{subsec:replan}

\begin{figure}[t]
	\vspace{-2ex}
	\centering
	\subfloat[Continuous replanning]{\label{fig:bad_replanning}\includegraphics[width=0.24\textwidth]{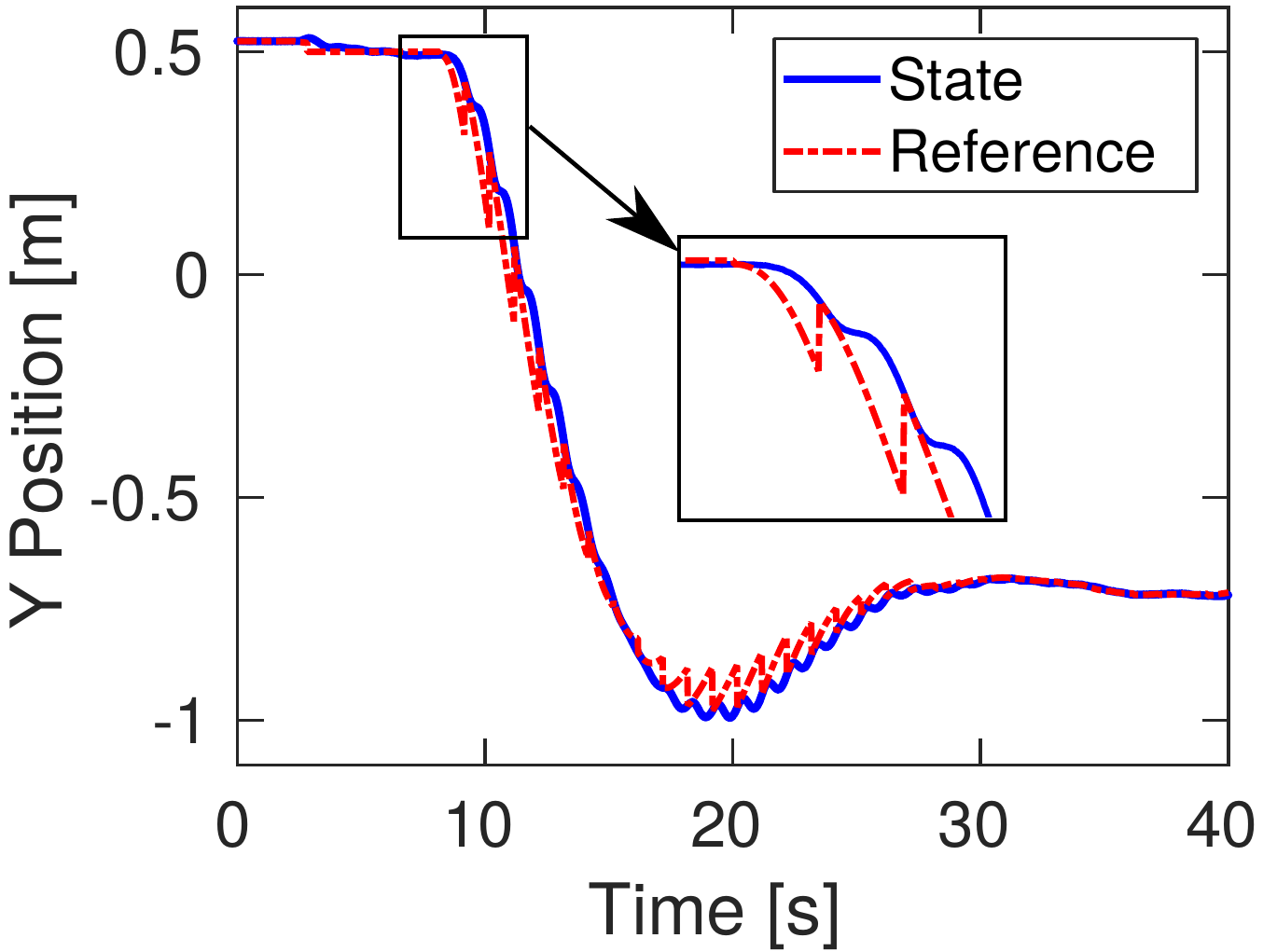}}
	\subfloat[Event-triggered replanning]{\label{fig:replanning}\includegraphics[width=0.23\textwidth]{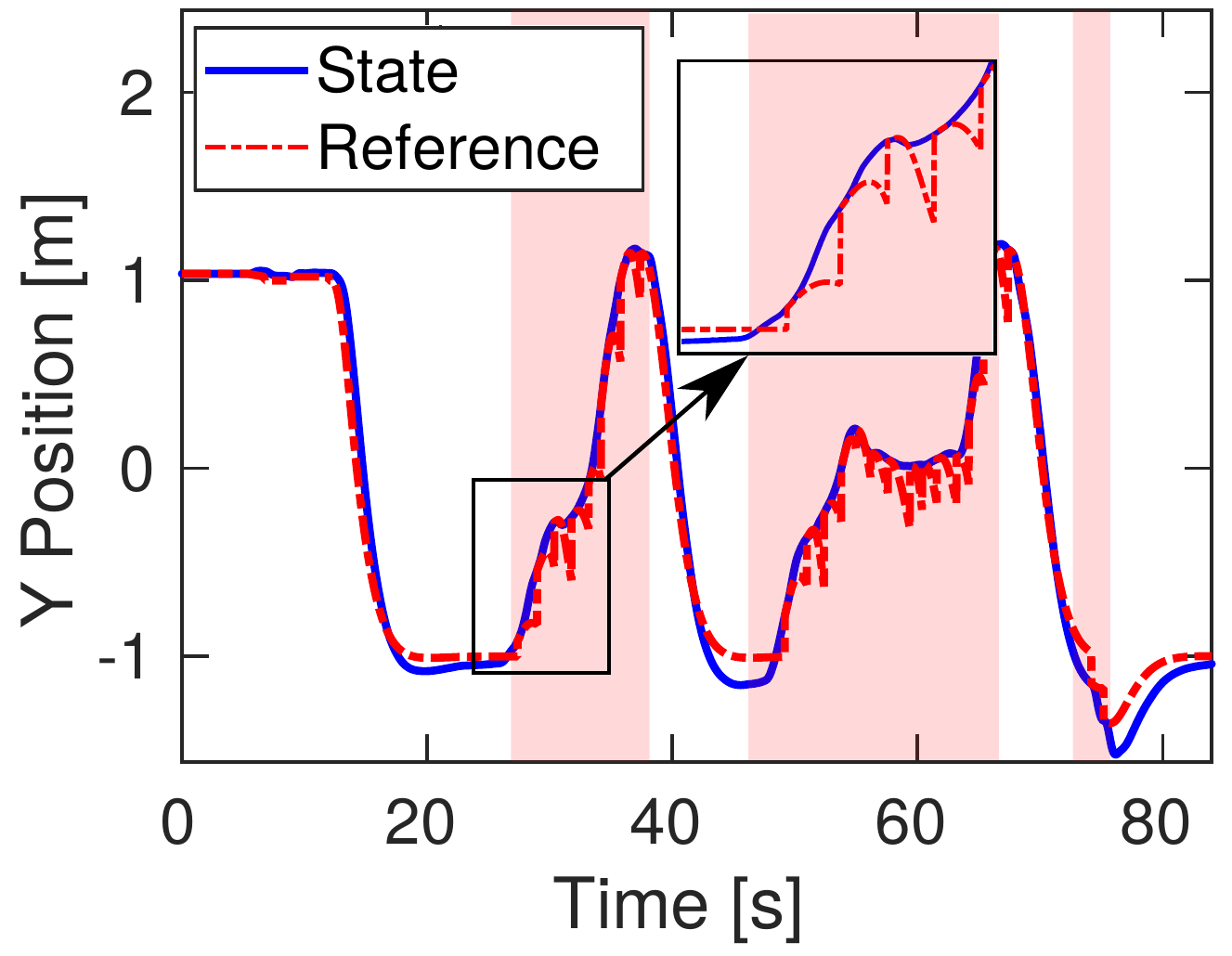}}
	\hfill
	\caption{Experimental data comparison of replanning strategies. Continuous replanning creates discontinuities in the reference signals that cause jittering in the state of the agent. The event-triggered strategy remedies this behaviour by introducing discontinuities only when the agents are being perturbed.}
	\label{fig:strategies}
\end{figure}

Choosing the initial condition for the input at each planning cycle to be equal to the current state of the robot was proposed in \cite{csenbacslar2019robust}, but it has certain limitations. First, if we require $C^r$-continuity on the inputs, then we need to reliably measure the $r^{th}$ derivative of the robot's position. Second, for imperfect trajectory tracking or systems with slow dynamics, this replanning strategy consistently causes (potentially big) discontinuities of the input to match the state of the robot, as shown in Fig.~\ref{fig:bad_replanning}. Such discontinuities cause undesired jittering in the robot and slow down its progress to complete the task.


To address these concerns, we propose an event-triggered replanning strategy, in which we reset the input to match the states of the agent only whenever we detect the agent has been perturbed. To detect such an event, we heuristically designed an activation function that serves to detect disturbances to the agent. An example of such an activation function for second-order tracking dynamics is:
\begin{equation}
\label{eq:activation}
f_n[k_t] = \frac{(\text{p}_{i,n}[k_t] - \text{u}_{i,n}[k_t])^5}{-(\text{v}_{i,n}[k_t] + sgn(\text{v}_{i,n}[k_t])\varepsilon)}, \quad n = 1,2,3
\end{equation}
where the subscript $n$ represents the spatial component ([$x,y,z$]) of the vectors associated with agent $i$. The term $(\text{p}_{i,n}[k_t] - \text{u}_{i,n}[k_t])$ is the trajectory tracking error, and the term $sgn(\text{v}_{i,n}[k_t])\varepsilon$ with a small scalar $\varepsilon \ll 1$ is used to avoid singularities in ${f}_n[k_t]$. We assume $|\text{v}_{i,n}[k_t]| > 0$, which is realistic in real-world operation due to noise in state estimation.


The intuition behind (\ref{eq:activation}) is that we want to reset our reference signal whenever the tracking error grows large. However, designing an appropriate threshold value for the tracking error is tricky due to its high variability during execution. Instead, ${f}_n[k_t]$ is designed to detect whenever the error is growing but the velocity is either small or growing in the opposite direction of the error. To detect these scenarios, we define the robot is operating normally if the inequality
\begin{equation}
{f}_{\min} < {f}_n[k_t] < {f}_{\max}
\label{eq:bounds}
\end{equation}
holds for every element of ${f}_n[k_t]$. The values of ${f}_{\min}$ and ${f}_{\max}$ must be chosen by extracting the extrema of ${f}_n[k_t]$ under normal operation. If (\ref{eq:bounds}) does not hold, then the agent is being disturbed and we set the initial position and velocity of the Bézier curve to match the states of the vehicle, while setting higher-order derivatives to zero. In summary,
\begin{equation}
	\textbf{u}_{0,i}[k_t] = \begin{cases}
	\bhat{\textbf{u}}_i[1|k_t - 1], & \text{if ${f}_{\min} < {f}_n[k_t] < {f}_{\max}$,} \\
	(\bar{\textbf{x}}_i[k_t], \bm{0}), & \text{else.}
	\end{cases}
\end{equation}
In order to validate the proposed replanning strategy, we conducted an experiment with our quadrotor platform while a human operator perturbed it along its path. The task of the quadrotor was to reach a y-coordinate of -1 m. The reference signal and state of the quadrotor are shown in Fig.~\ref{fig:replanning}, where the red segments mean the agent was being disturbed. During these disturbed stages we observe how the reference signal is frequently reset to match the state of the robot. The replanning helps the quadrotor continue its task whenever the disturbance is removed. Under normal operation (white segments) the replanning is not required, which leads to a smooth reference signal that avoids the shortcomings observed in Fig.~\ref{fig:bad_replanning}. \todo{Note that we assume that the perturbations not detected by the activation function (\ref{eq:activation}) can be rejected by the underlying controller, which we validated experimentally.} The video that accompanies this paper showcases the strategy working in experiments with quadrotors. \todo{These disturbances would have led to crashing and mission failure with typical offline approaches (e.g., \cite{luis2019trajectory}), since there is no adaptation of the pre-planned reference signal, and the underlying control system would have been unable to reject the perturbations.}

\section{The Algorithm}
\label{sec:algorithm}

\begin{algorithm}[t]
	\label{alg:update}
	\SetKwData{Left}{left}\SetKwData{This}{this}\SetKwData{Up}{up}
	\SetKwFunction{Union}{Union}\SetKwFunction{FindCompress}{FindCompress}
	\SetKwInOut{Input}{Input}\SetKwInOut{Output}{Output}
	\SetKwFunction{setTargetLocation}{setTargetLocation}
	\SetKwFunction{getInitialReference}{getInitRef}
	\SetKwFunction{getCollisionTuple}{getCollision}
	\SetKwFunction{buildQP}{buildQP}
	\SetKwFunction{solve}{solve}
	\SetKwFunction{updateHorizon}{\todo{broadcastUpdatedHorizon}}
	\SetKwFunction{updateInitialReference}{updateInitialReference}
	\SetKwFunction{getNextInputs}{getSampledInput}
	\SetKwFunction{receiveAgentPredictions}{receiveAgentPredictions}
	
	\Input {\todo{Current states of all agents ($\textbf{x}[k_t]$), target location ($\textbf{p}_{d,i}$)}}
	\Output {Commands to be applied from $t_0$ to $t_0 + h$ with sampling of $T_s$ (\todo{$\bar{\textbf{u}}_i$)}}
	
	\setTargetLocation($\textbf{p}_{d,i}$)
	
	\todo{$\bm{\Pi}[k_t-1]$ $\leftarrow$ \receiveAgentPredictions()}

	$\textbf{u}_{0,i}[k_t]$ $\leftarrow$ \getInitialReference($\bar{\textbf{x}}_i[k_t]$, $\bhat{\textbf{u}}_i[1|k_t - 1]$)
	
	($\textbf{A}_{\text{coll}},\, \textbf{b}_{\text{coll}}$) $\leftarrow$ \getCollisionTuple($\bar{\textbf{x}}[k_t]$, $\bm{\Pi}[k_t-1]$)
	
	QP $\leftarrow$  \buildQP($\textbf{A}_{\text{coll}},\, \textbf{b}_{\text{coll}}$, $\textbf{u}_{0,i}[k_t]$, $\bar{\textbf{x}}_i[k_t]$)
	
	$\bm{\mathcal{U}}_i$ $\leftarrow$ \solve(QP)
	
	$\bm{\Pi}_i[k_t]$ $\leftarrow$ \updateHorizon($\bm{\mathcal{U}}_i$, $\bar{\textbf{x}}_i[k_t]$)
	
	$\bhat{\textbf{u}}_i[1|k_t]$ $\leftarrow$ \updateInitialReference($\bm{\mathcal{U}}_i$)
	
	$\bar{\textbf{u}}_i$ $\leftarrow$ \getNextInputs($\bm{\mathcal{U}}_i$)
	
	\KwRet {\todo{\normalfont$\bar{\textbf{u}}_i$}}
	\caption{Input updates for agent $i$}
\end{algorithm}
In this section we describe the core algorithm used to update the optimal input sequence for each agent, outlined in Alg.~\ref{alg:update}. \todo{As stated, the algorithm is conceived to be executed in a distributed fashion by a group of agents with communication capabilities}. It takes as inputs the measured state of each agent and the desired location of agent $i$. For execution we consider two different time bases: one with a coarse time step $h$, used for the MPC planning, and one with a refined time step $T_s$ used for commanding the agents at a higher rate. With this definition, the output of Alg.~\ref{alg:update} is the set of inputs for agent $i$ in the time frame in-between planning cycles, i.e., $t \in [t_0, \, t_0 + h]$ with sample rate $T_s$. In other words, the output are subsamples of the input between $\bhat{\textbf{u}}_i[0|k_t]$ and $\bhat{\textbf{u}}_i[1|k_t]$. \todo{This subsampling process is exact and not an approximation, due to the chosen continuous parameterization of trajectories.}

In line 1 we build the error penalties given by (\ref{eq:error_cost}), which is only required if the setpoint $\textbf{p}_{d,i}$ of the agent changes. \todo{In line 2 each agent receives the latest predictions of all the other agents through some communication channel, which is considered ideal in this paper, with no delays or packet drops. In lines 3-9 we update the input sequence of each agent. Note that the first time the algorithm is executed, the latest predictions may be initialized trivially by assuming static neighbours}. First, in line 3 we apply the event-triggered replanning strategy to decide the value of the initial condition of the input. The collision avoidance constraint (BVC or on-demand) is constructed in line 4. Note that BVC would not require the prediction information, \todo{meaning that there is no communication between agents}, but instead it would require the measured state of the agents. Conversely, on-demand avoidance only requires the predictions and not the measured states. Lines 5-6 build and solve the standard quadratic programming problem outlined in (\ref{eqn:convex_coll}). Once the solution vector $\bm{\mathcal{U}}_i$ is obtained from the QP solver, we can then sample the resulting Bézier curves to obtain a sampled representation of the input (or the state). \todo{Line 7 updates the horizon for the agent and broadcasts the information to the rest of the team}. Line 8 updates the initial condition of the reference to be used in the next planning cycle, in the case where replanning is not required. Lastly, line 9 samples the resulting Bézier curve with period $T_s$ to obtain the sequence to be applied for $t \in [t_0, \, t_0 + h]$.

\todo{Since our physical platform does not have inter-agent communication capabilities, Alg.~\ref{alg:update} is implemented in a centralized manner from an offboard computer. The centralized implementation of Alg.~\ref{alg:update} is executed in parallel for all the agents, since there is no inter-agent data dependency for the input updates.}
\section{Simulation Results}
\label{sec:simulation}

We created a simulation environment in MATLAB 2017a and executed it on a PC with Intel Xeon CPU with 8 cores and 16 GB of RAM, running at 3 GHz. The agents were modeled after the Crazyflie 2.0 quadrotor, using ${r_{\min} = 0.3 \, \text{m}}$ and $\bm{\Theta} = \text{diag}([1, 1, 2])$, but a collision was declared using $r_{\text{coll}} = 0.2 \, \text{m}$ (closer to the physical size of the quadrotor) and $\bm{\Theta}_{\text{coll}} = \text{diag}([1, 1, 2.25])$. The trajectory tracking dynamics were identified by fitting a second-order model to experimental data from the step response of the system depicted in Fig.~\ref{fig:block}. We selected a step of $h = 0.2 \, \text{s}$, which means that trajectories are replanned at only 5 Hz.

For the input sequence we chose Bézier curves with $p=5$, $l = 3$ and $t_h = 3 \, \text{s}$, where each segment had a fixed duration of 1 second. Additionally, we imposed actuation limits with $\gamma^{(2)}_{\max} = -\gamma^{(2)}_{\min} = 1 \, \text{m/s}^2$. After tuning the cost function, we selected $\kappa = 3$, $q_k = 100$, $\alpha_2 = 0.008$, $\zeta = 1$ and $\xi = -5\times10^4$. For the replanning function in (\ref{eq:activation}) we chose $\epsilon = 0.01$, $f_{\min} = -0.01$, and $f_{\max} = 0.8$. We also added noise in the measured state $\bar{\textbf{x}}_i[k_t]$ based on empirical data gathered from an overhead motion capture system.

\subsection{Comparison of Collision Avoidance Methods}
\label{subsec:sim_comp}
We compared four different optimization-based collision avoidance methods in random transition scenarios: 1) BVC as proposed in \cite{csenbacslar2019robust} (without the discrete planner component), 2) BVC using soft constraints, 3) On-demand collision avoidance applied in the state space and 4) On-demand collision avoidance in the input space. \todo{Discrete planning was removed for comparison purposes, but it should be noted that, in general, they can improve the performance of any of the methods. For instance, discrete planners may provide intermediate goal points for the agents to complete the transition task, fitting seamlessly with our current setup.}

\begin{figure}[t]
	\vspace{-2ex}
	\centering
	\subfloat[]{\label{fig:prob}\includegraphics[width=0.24\textwidth]{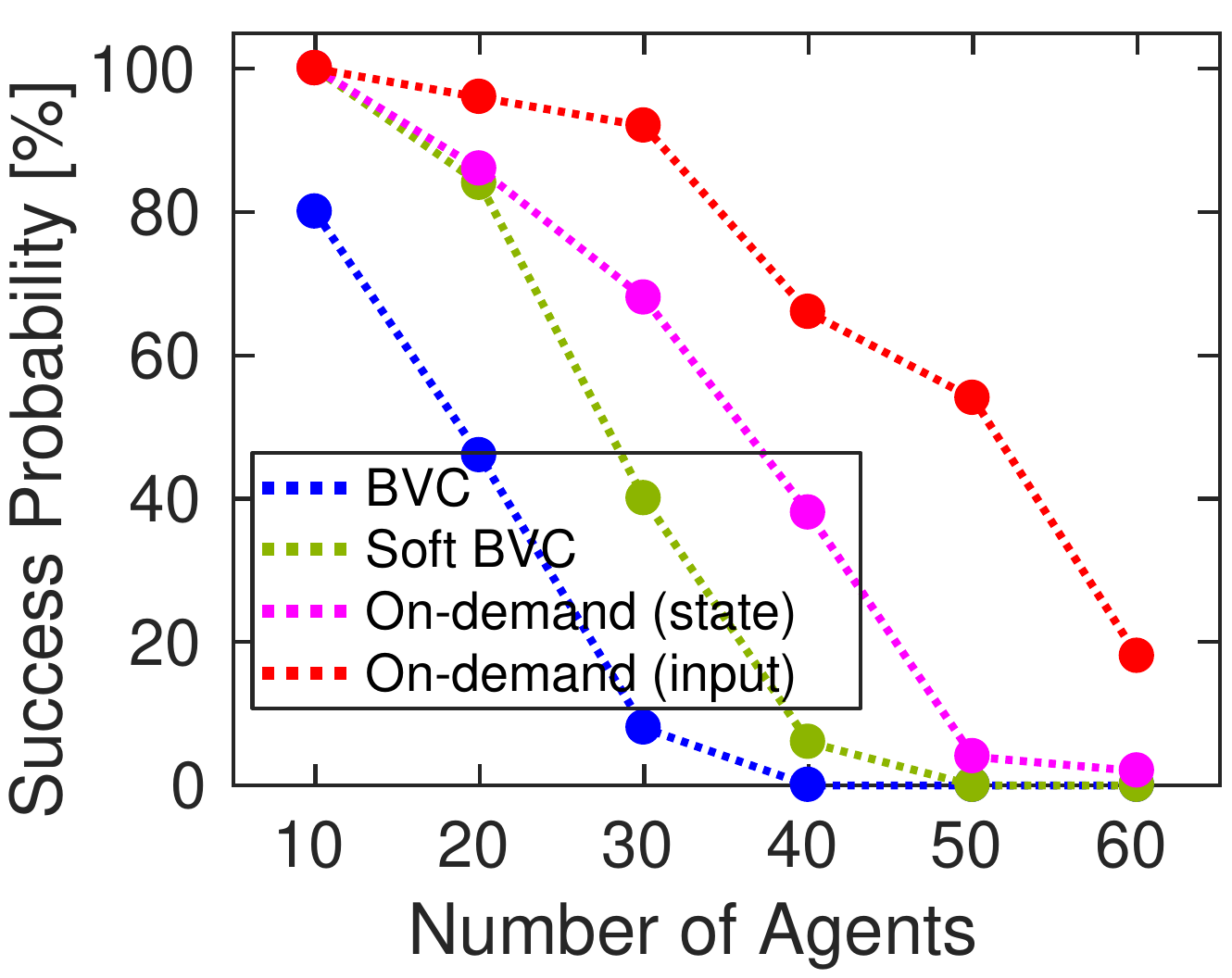}}
	\subfloat[]{\label{fig:transit_time}\includegraphics[width=0.24\textwidth]{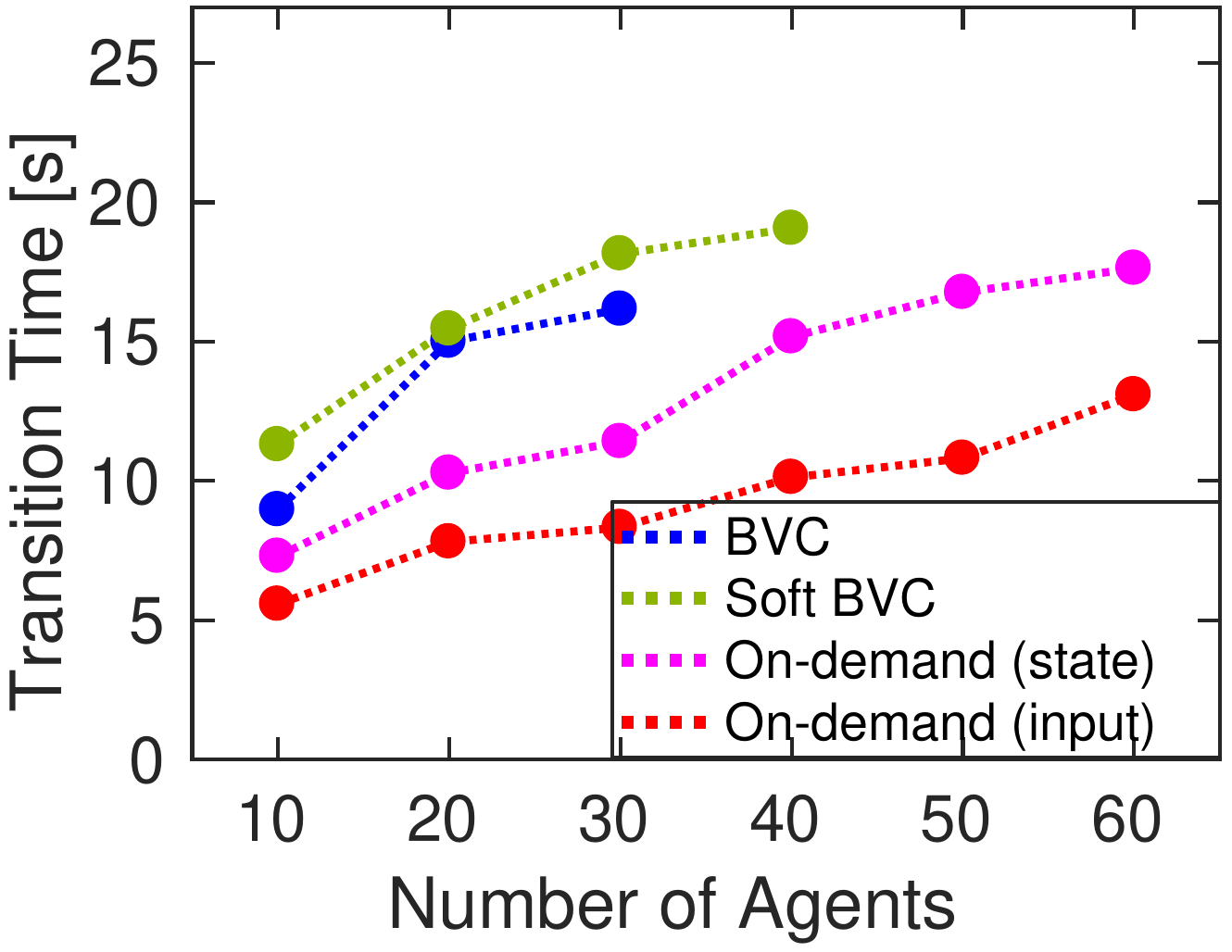}}
	\hfill
	\caption{Simulation performance comparison of various collision avoidance strategies. We considered different numbers of agents in a fixed volume of $18 \hspace{1ex} \text{m}^3$. For each swarm size, 50 different random test cases were generated and averaged.}
	\label{fig:perf}
\end{figure}

\begin{table*}[t]
	\centering
	\vspace{1ex}
	\caption{Experimental results summary for random transition tasks involving increasing number of agents.}
	\begin{tabular}{l|l|l|l|l|l|l|l|l|l|l}
		\toprule
		\# Agents & 2 & 4 & 6 & 8 & 10 & 12 & 14 & 16 & 18 & 20 \\
		\midrule
		Avg. solve time of Alg. 1 [ms] & 3.3 & 6.2 & 9.9 & 13.8 & 16.9 & 17.6 & 20.3 & 20.5 & 23.4 & 28.3\\
		Std. solve time of Alg. 1 [ms] & 0.4 & 0.9 & 1.7 & 3.3 & 5.1 & 7.6 & 8.0 & 8.9 & 10.2 & 11.4 \\
		Min. distance [cm] & 36.2 & 32.2 & 31.1 & 30.0 & 29.2 & 28.6 & 29.1 & 26.0 & 26.1 & 25.3 \\
		\bottomrule
	\end{tabular}
	\label{tab:summary}
\end{table*}

We considered a fixed-volume, obstacle-free workspace of $18 \hspace{1ex} \text{m}^3$ (roughly the size of our indoor flight arena), with randomly generated initial and final locations for all agents. The number of agents varied from 10 to 60, in order to test the algorithms as the agent density increased.  A trial was considered successful if all agents were able to reach their goals without collisions and within 20 seconds. After each simulation we ran a collision check (using $r_{\text{coll}}$ and $\bm{\Theta}_{\text{coll}}$) and a goal check (allowing 10 cm distance from the target location) to determine if the test was successful.

In Fig.~\ref{fig:perf} we show the performance obtained using each method. The success probability for each swarm size considered is highlighted in Fig.~\ref{fig:prob}. We notice that as the number of agents increases (ergo, a denser workspace) the effectiveness of the BVC methods decay drastically. Using soft constraints helps, but ultimately the approach is too conservative to resolve transition scenarios with a high density of agents.

On the other hand, the on-demand collision avoidance strategy shows better performance when applied in the input space, especially in high agent density workspaces. Input-space collision avoidance achieved more than 90\% success rate with swarm sizes up to 30 agents. We observe a significant decline in performance after 30 agents in all the tested methods. This is a weakness of our approach given the need to relax collision constraints in order to find solutions. As the density grows, then higher relaxations will be required to solve the transitions, which may result in collisions.

One explanation to the performance difference between state and input space avoidance resides on the agent model. In the identified dynamics, the position of the agents is, essentially, a delayed version of the input signal (with some overshoot). Thus, by treating collision avoidance in the input space, the agents are preemptively avoiding each other, which ultimately leads to less collisions during execution. Also, by using the inputs as opposed to predicted states, the collision avoidance is less sensitive to the model's accuracy. \todo{We note that for other linear systems where the state and the input are completely different quantities, performing input-space collision avoidance is not guaranteed to be safe (since non-colliding inputs do not translate into non-colliding positions) and state-space avoidance should be preferred.}

In the same order of ideas, Fig.~\ref{fig:transit_time} shows that, on average, using on-demand collision avoidance leads to faster transition times than the BVC methods, averaging around 50\% transition time reductions. These numbers match the analysis made on Sec.~\ref{subsec: avoidance} using Fig.~\ref{fig:collision_avoidance}. \todo{We also analyzed the average travelled distance by all the agents as a measure of optimality. The simulation data suggested that all the methods produce trajectories of almost the same length, with a slight advantage to the soft BVC method which produces, on average, trajectories around 3\% shorter than the rest.}

\subsection{Runtime Benchmark}
\begin{figure}[t]
	\centering
	\includegraphics[width=0.5\columnwidth]{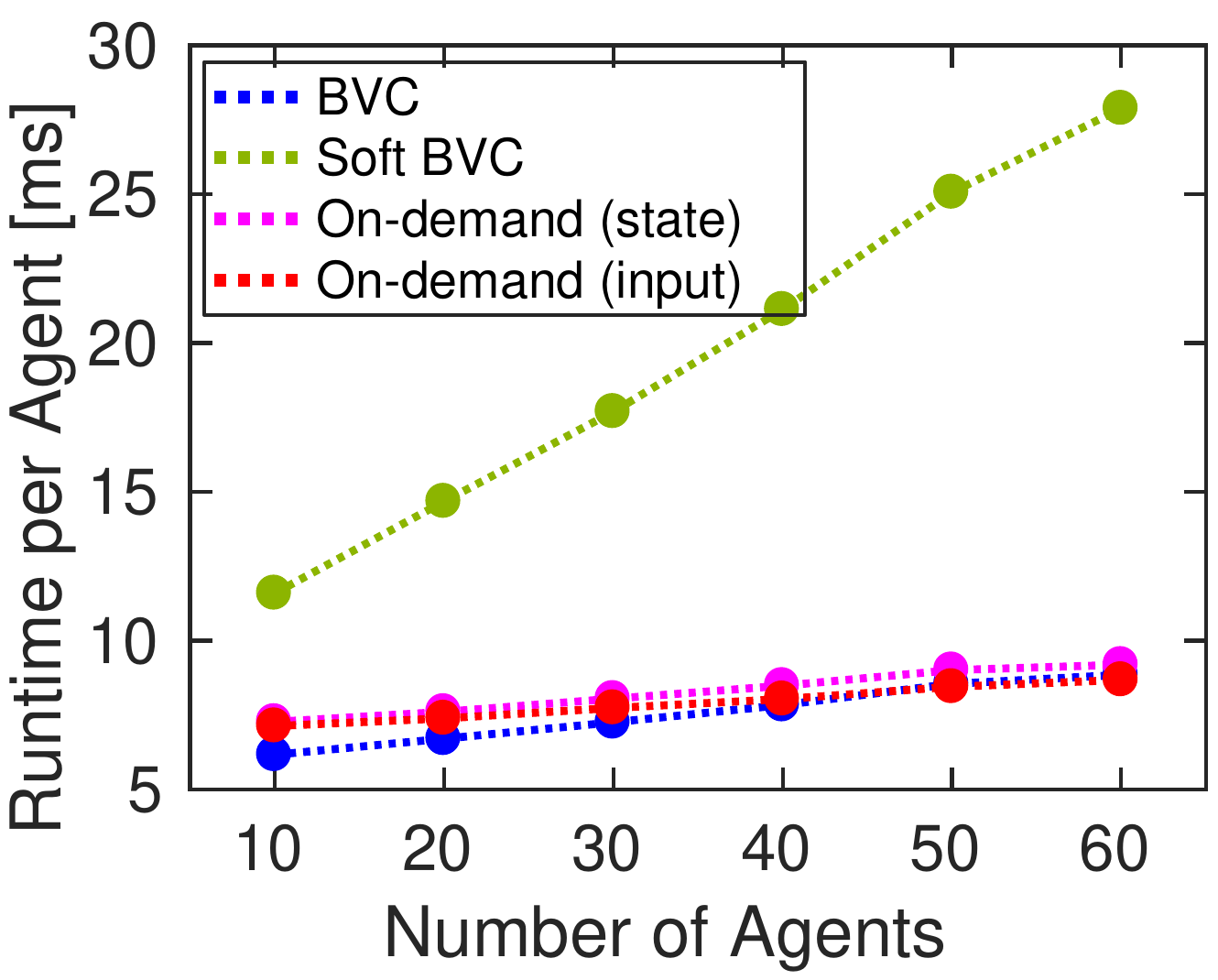}
	\caption{Comparison of the average runtime per agent to update the inputs using our on-demand collision avoidance and the BVC method. The data shown is the average over 50 randomly generated tests for each swarm size considered. 
	}
	\label{fig:computation}
\end{figure}

We compared the computation time per agent to update their input sequence. In Fig.~\ref{fig:computation} the results are presented, where we specifically show the average time per agent to solve the associated QP problem.

To formally analyze the scaling of both algorithms, define $N_{i,k_t}$ to be the number of nearby neighbours of agent $i$ to be considered for collision avoidance at time step $k_t$. The amount of inequality constraints on both BVC and on-demand methods scale with $\mathcal{O}(N_{i,k_t})$. The soft BVC method adds additional $(p + 1)\times N_{i,k_t}$ slack variables to relax the constraints, while the on-demand methods add only $N_{i,k_t}$ new decision variables to the problem. In Fig.~\ref{fig:computation} we observe the empirical runtime of the considered methods. The soft BVC method has the slowest runtime, due to the added slack variables and overall bigger problems to be solved. For the other three methods the runtime is fairly similar, with a slight advantage to the BVC method. 

\section{Experimental Results}
\label{sec:experiments}

The online generation method outlined in this paper (with on-demand avoidance) was implemented in C++, using ROS to manage the drone swarm and qpOASES \cite{Ferreau2014} as the QP solver. In this section we provide experimental results using our Crazyflie 2.0 swarm testbed. All the inputs were computed from a single computer and broadcast to the swarm through a radiolink, alongside the estimated position of each individual agent given by a motion capture system. The computer specs and algorithm parameters are the same as Sec.~\ref{sec:simulation}, with the exception of $\xi = -1\times10^3$ and the addition of $T_s = 0.05 \, \text{s}$ (trajectories were being sent to the swarm at $20 \, \text{Hz}$).


A video summarizing the experimental results can be found at \href{http://tiny.cc/online-dmpc}{{\tt http://tiny.cc/online-dmpc}}.

\subsection{Obstacle-Free Transitions}
We considered different swarm sizes, ranging from 2 to 20 drones. For each swarm size, three independent flights were executed, where each flight consisted in 30 seconds of randomly generated transitions. The agents were restricted to move in a $3\times 3\times 2\, \text{m}^3$ volume, and we used $r_{\min} = 0.35 \, \text{m}$ as the safety distance.

For each test, we recorded the average computation time to execute Alg.~1 and the minimum inter-agent distance during the flight. The results are summarized in Table~\ref{tab:summary}. As expected, the average computation time increases as we add more agents to the problem, since all computations are being executed by a single computer. The interesting result is that the scaling we obtain in runtime is pseudo-linear, since we are able to parallelize the computation thanks to the distributed nature of the approach. 

\begin{figure}[t]
	\centering
	\includegraphics[width=0.7\columnwidth]{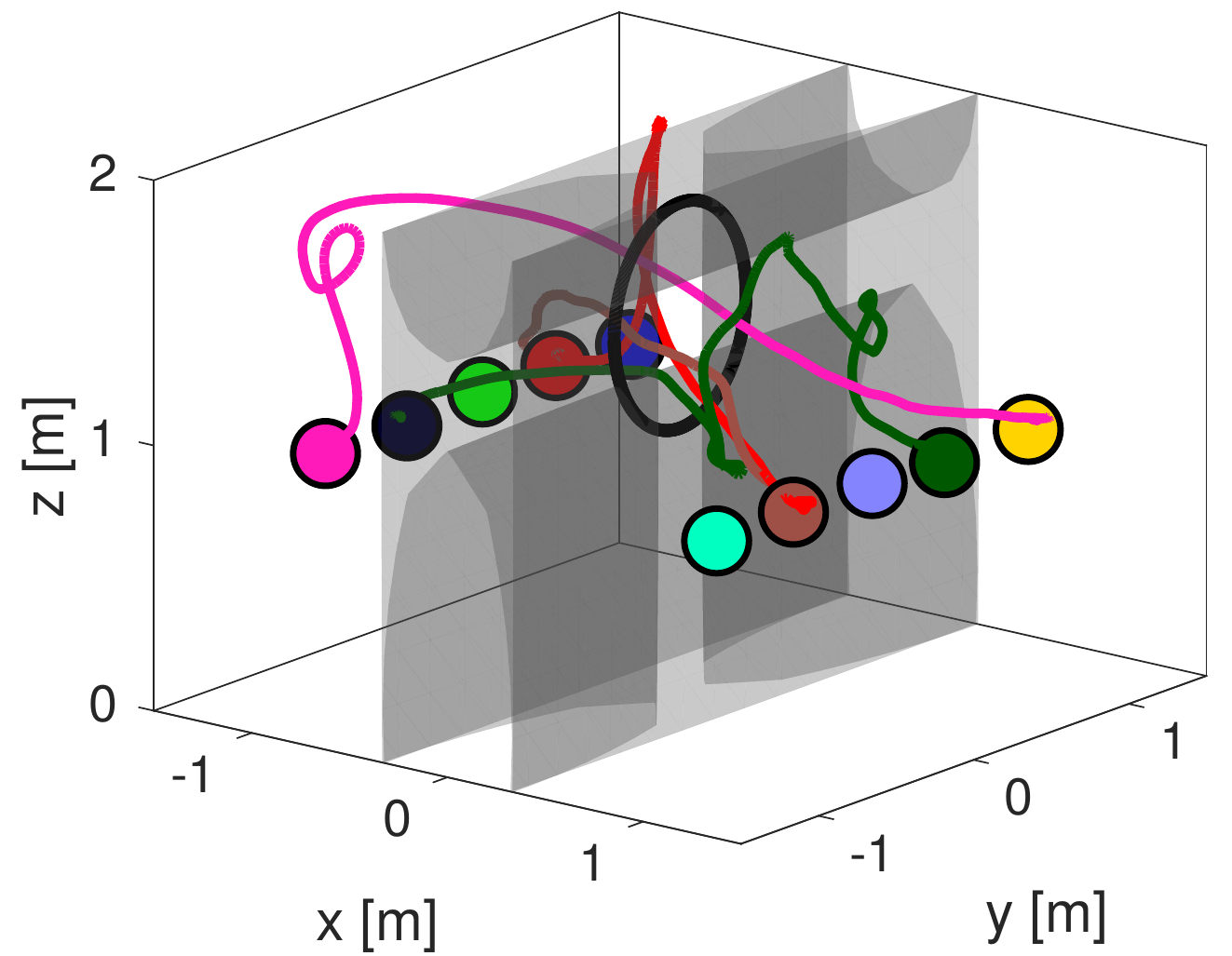}
	\caption{A 10-drone transition passing through a hula-hoop (denoted by the black circle). The forbidden space is defined by four ellipsoids acting as static obstacles. The coloured circles denote the initial locations of the agents, and the corresponding coloured lines are the followed trajectories towards the antipodal goal locations (showing only four trajectories for clarity). }
	\label{fig:hoop}
\end{figure}

The minimum inter-agent distance decreases as we increase the number of agents, i.e., there is less available space to move collision-free. Since the optimizer is allowed to violate the collision constraint, the original margin of $r_{\min} = 0.35 \, \text{m}$ is violated if required. Such scenarios of violation appear more often the higher the agent density in the workspace. Although this is suboptimal from a safety perspective, experiments show that as long as a sufficiently large $r_{\min}$ is chosen, the amount of violation incurred while optimizing will still allow the agents to move collision-free.

\subsection{Transition Tasks with Static Obstacles}

We tasked a group of drones to exchange positions with each other by passing through a hula-hoop with a 85 cm diameter. The environment was divided by an invisible wall with a passage-way defined by the hula-hoop. In Fig.~\ref{fig:hoop} we show the 10-drone transition scenario solved in experiments. Note that static obstacles are added to the problem as new ``neighbours'' for each agent, with their own ellipsoidal parameters, which means that the runtime complexity scales linearly with the number of static obstacles.

\begin{figure}[t]
	\centering
	\includegraphics[width=0.85\columnwidth]{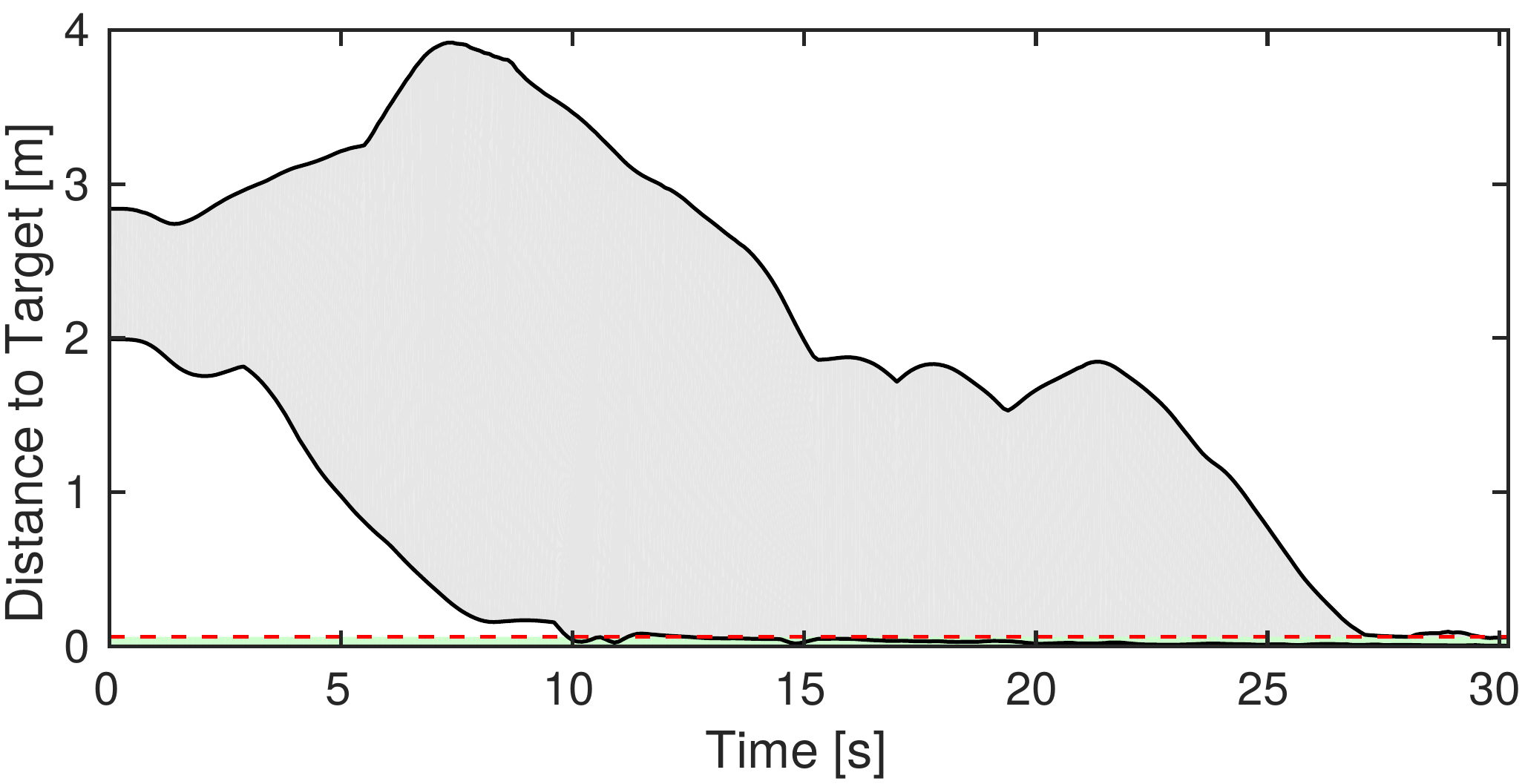}
	\caption{Distance to target envelope (minimum and maximum over time) of the 10-drone hula-hoop transition task. The light green section near the bottom represents the zone where transition success is declared: a 6 cm radius of the target location. In this case the transition was completed in $T_f = 28 \, \text{s}$.}
	\label{fig:target}
\end{figure}

The restricted zone was modeled as the union of four ellipsoids. They are shaped in such a way that they are intersecting and provide a small gap of $30\times 30\, \text{cm}^2$ for the agents to pass through. In the tracked trajectories we observe that some of the agents were able to fly directly through the circle, while others took detours in order to let other agents pass first. The distance-to-goal envelope shown in Fig.~\ref{fig:target} demonstrates how the agents make progress over time to decrease the distance towards their goal, eventually converging to it within some tolerance region. 



Several different tuning parameters were tried while solving this particular task. While using input space collision avoidance, a wide range of penalty gains and maximum accelerations worked well to solve the task; \todo{the completion time $T_f$} varied from 20.1 to 48.4 seconds in 18 different trials, with \todo{$\xi \in [-1.5, -0.5]\times10^3$}. \todo{In most cases, a penalty of $\xi < -2\times10^3$ resulted in oscillatory behaviour near the obstacles, which sometimes led to deadlocks. On the other hand, the success rate using state space collision avoidance was much lower, resulting in deadlocks much more frequently than with input-space avoidance.}

\todo{Non-convex obstacles could lead to agents getting stuck in local minima, in which case the use of discrete planners to provide intermediate waypoints may be required.} 


\section{Conclusion and Future Work}
\label{sec:conclusion}
In this paper we presented a framework for multi-robot online trajectory generation based on distributed model predictive control (DMPC). In transition tasks, our method has a higher success rate and lower travel times than using the Buffered Voronoi Cells method. The simulations indicated more than 90\% success rate with up to 30 palm-sized quadrotor agents in a $18 \hspace{1ex} \text{m}^3$ arena.

The parallelization of the method leads to high scalability. In experiments we were able to send trajectories in real-time (20 Hz) for a swarm of 20 quadrotors. Our approach showed satisfactory results in a complicated transition scenario passing through a hula-hoop, and robust replanning in the presence of unmodeled disturbances.

\todo{One interesting area of future work includes accommodating a more realistic communication channel in the formulation, in order to deal with delays and packet drops.}

\bibliographystyle{IEEEtran}
\bibliography{IEEEabrv,reference}

\begin{thebibliography}{10}
\providecommand{\url}[1]{#1}
\csname url@samestyle\endcsname
\providecommand{\newblock}{\relax}
\providecommand{\bibinfo}[2]{#2}
\providecommand{\BIBentrySTDinterwordspacing}{\spaceskip=0pt\relax}
\providecommand{\BIBentryALTinterwordstretchfactor}{4}
\providecommand{\BIBentryALTinterwordspacing}{\spaceskip=\fontdimen2\font plus
\BIBentryALTinterwordstretchfactor\fontdimen3\font minus
  \fontdimen4\font\relax}
\providecommand{\BIBforeignlanguage}[2]{{%
\expandafter\ifx\csname l@#1\endcsname\relax
\typeout{** WARNING: IEEEtran.bst: No hyphenation pattern has been}%
\typeout{** loaded for the language `#1'. Using the pattern for}%
\typeout{** the default language instead.}%
\else
\language=\csname l@#1\endcsname
\fi
#2}}
\providecommand{\BIBdecl}{\relax}
\BIBdecl

\bibitem{augugliaro2012generation}
F.~Augugliaro, A.~P. Schoellig, and R.~D'Andrea, ``Generation of collision-free
  trajectories for a quadrocopter fleet: A sequential convex programming
  approach,'' in \emph{IEEE/RSJ International Conference on Intelligent Robots
  and Systems (IROS)}, 2012, pp. 1917--1922.

\bibitem{chen2015decoupled}
Y.~Chen, M.~Cutler, and J.~P. How, ``Decoupled multiagent path planning via
  incremental sequential convex programming,'' in \emph{IEEE International
  Conference on Robotics and Automation (ICRA)}, 2015, pp. 5954--5961.

\bibitem{van2017distributed}
R.~Van~Parys and G.~Pipeleers, ``Distributed model predictive formation control
  with inter-vehicle collision avoidance,'' in \emph{Asian Control Conference
  (ACC)}, 2017.

\bibitem{luis2019trajectory}
C.~E. Luis and A.~P. Schoellig, ``Trajectory generation for multiagent
  point-to-point transitions via distributed model predictive control,''
  \emph{IEEE Robotics and Automation Letters}, vol.~4, no.~2, pp. 375--382,
  2019.

\bibitem{vcap2013multi}
M.~{\v{C}}{\'a}p, P.~Nov{\'a}k, J.~Vokr{\'\i}nek, and M.~P{\v{e}}chou{\v{c}}ek,
  ``Multi-agent {RRT}: sampling-based cooperative pathfinding,'' in
  \emph{Proceedings of the International Conference on Autonomous Agents and
  Multi-agent Systems}, 2013, pp. 1263--1264.

\bibitem{honig2018trajectory}
W.~H{\"o}nig, J.~A. Preiss, T.~S. Kumar, G.~S. Sukhatme, and N.~Ayanian,
  ``Trajectory planning for quadrotor swarms,'' \emph{IEEE Transactions on
  Robotics}, vol.~34, no.~4, pp. 856--869, 2018.

\bibitem{hamer2018fast}
M.~Hamer, L.~Widmer, and R.~D’Andrea, ``Fast generation of collision-free
  trajectories for robot swarms using {GPU} acceleration,'' \emph{IEEE Access},
  vol.~7, pp. 6679--6690, 2018.

\bibitem{van2011reciprocal}
J.~Van Den~Berg, J.~Snape, S.~J. Guy, and D.~Manocha, ``Reciprocal collision
  avoidance with acceleration-velocity obstacles,'' in \emph{International
  Conference on Robotics and Automation}, 2011, pp. 3475--3482.

\bibitem{alonso2018cooperative}
J.~Alonso-Mora, P.~Beardsley, and R.~Siegwart, ``Cooperative collision
  avoidance for nonholonomic robots,'' \emph{IEEE Transactions on Robotics},
  vol.~34, no.~2, pp. 404--420, 2018.

\bibitem{zhou2017fast}
D.~Zhou, Z.~Wang, S.~Bandyopadhyay, and M.~Schwager, ``Fast, on-line collision
  avoidance for dynamic vehicles using buffered voronoi cells,'' \emph{IEEE
  Robotics and Automation Letters}, vol.~2, no.~2, pp. 1047--1054, 2017.

\bibitem{csenbacslar2019robust}
B.~{\c{S}}enba{\c{s}}lar, W.~H{\"o}nig, and N.~Ayanian, ``Robust trajectory
  execution for multi-robot teams using distributed real-time replanning,'' in
  \emph{Distributed Autonomous Robotic Systems}.\hskip 1em plus 0.5em minus
  0.4em\relax Springer, 2019, pp. 167--181.

\bibitem{hernandez2016distributed}
B.~Hernandez and P.~Trodden, ``Distributed model predictive control using a
  chain of tubes,'' in \emph{UKACC 11th International Conference on Control
  (CONTROL)}, 2016, pp. 1--6.

\bibitem{nikou2018decentralized}
A.~Nikou and D.~V. Dimarogonas, ``Decentralized tube-based model predictive
  control of uncertain nonlinear multiagent systems,'' \emph{International
  Journal of Robust and Nonlinear Control}, vol.~29, no.~10, pp. 2799--2818,
  2019.

\bibitem{mellinger2011minimum}
D.~Mellinger and V.~Kumar, ``Minimum snap trajectory generation and control for
  quadrotors,'' in \emph{IEEE International Conference on Robotics and
  Automation (ICRA)}, 2011, pp. 2520--2525.

\bibitem{joy2000bernstein}
K.~I. Joy, ``Bernstein polynomials,'' \emph{On-Line Geometric Modeling Notes},
  vol.~13, 2000.

\bibitem{mercy2017spline}
T.~Mercy, R.~Van~Parys, and G.~Pipeleers, ``Spline-based motion planning for
  autonomous guided vehicles in a dynamic environment,'' \emph{IEEE
  Transactions on Control Systems Technology}, no.~99, pp. 1--8, 2017.

\bibitem{charles2013polynomial}
R.~Charles, A.~Bry, and N.~Roy, ``Polynomial trajectory planning for quadrotor
  flight,'' in \emph{IEEE International Conference on Robotics and Automation
  (ICRA)}, 2013.

\bibitem{Ferreau2014}
H.~Ferreau, C.~Kirches, A.~Potschka, H.~Bock, and M.~Diehl, ``{qpOASES}: A
  parametric active-set algorithm for quadratic programming,''
  \emph{Mathematical Programming Computation}, vol.~6, no.~4, pp. 327--363,
  2014.

\end{thebibliography}

\end{document}